\documentclass[11pt,twocolumn]{article}
\usepackage[preprint]{acl}
\usepackage{longtable}
\usepackage{booktabs}
\usepackage[utf8]{inputenc}
\usepackage{times}
\usepackage{microtype}
\usepackage{array}
\usepackage{amsmath}
\usepackage{graphicx}
\usepackage{enumitem}
\usepackage{listings}
\usepackage[most]{tcolorbox}
\hypersetup{hypertexnames=false}
\usepackage{url}
\usepackage{float}

\setlist[itemize]{leftmargin=1em, topsep=2pt, itemsep=1pt, parsep=0pt}
\title{When Reasoning Hurts: Source-Aware Evaluation of Frontier LLMs for Clinical SOAP Note Generation}

\author{Faizan Faisal \\
  University of California, Davis \\
  \texttt{fznfaisal@ucdavis.edu}
}

\begin{document}
\maketitle

\begin{abstract}
Reasoning-enabled LLMs perform strongly on medical-reasoning benchmarks, but
it is unclear whether those gains transfer to structured clinical
documentation. We test this for SOAP note generation from clinical dialogue
using a source-aware benchmark spanning OMI Health, ACI-Bench, and PriMock57.
GPT-5.4, DeepSeek-V4-Flash, and Gemma-4-E4B are evaluated in a controlled
$2{\times}2$ design that independently toggles provider-native reasoning and
same-source retrieval-augmented generation (RAG). Outputs are scored with
seven automatic metrics and two reference-aware LLM judges. Both evaluation paradigms agree: a non-reasoning GPT-5.4 configuration gives
the best overall quality, while DeepSeek-V4-Flash leads among
reasoning-enabled runs. Enabling reasoning substantially penalises GPT-5.4
across all three datasets, whereas same-source RAG provides smaller,
model-dependent gains. These results show that stronger reasoning capability
should not be assumed to improve fidelity-sensitive SOAP note generation
without task-specific evaluation.
\end{abstract}

\section{Introduction}

Clinical documentation consumes a disproportionate share of clinician time,
and there is growing interest in systems that can automatically convert
patient--clinician conversations into structured notes
\citep{yim2023aci, neupane2025clinicsum, goyal2025specialtyscribe}.
SOAP note generation is particularly appealing because the four-section format
(Subjective, Objective, Assessment, Plan) is both clinically familiar and
directly usable in downstream documentation workflows.
The task is nevertheless demanding: a high-quality note must be concise,
complete, and strictly grounded in dialogue evidence---plausible-sounding
content that is not supported by the conversation is a patient-safety risk,
not merely an annotation error.

Frontier models now expose explicit reasoning modes, and recent medical
reasoning benchmarks report encouraging gains on diagnostic and treatment
tasks \citep{qiu2025medr}.
For clinical \emph{text} tasks, however, the picture is more nuanced.
Chain-of-thought prompting has been shown to \emph{degrade} performance on
real-world clinical text understanding \citep{wu2025cotfails}, and
evaluation-oriented frameworks such as MEDIVAL examine reasoning-enhanced
\emph{judging} of documentation rather than reasoning-enhanced
\emph{generation} of it \citep{choi2025medival}.
These observations raise two linked questions that we set out to answer:
(1) when the task is structured SOAP note generation, does provider-native
reasoning actually help, and (2) do reference-aware LLM judges support
the same conclusions as automatic overlap metrics?

To answer both questions, we construct a source-aware benchmark from three
complementary datasets and evaluate three frontier models under a fully
factorial $2{\times}2$ design that independently toggles reasoning and RAG.
Using per-source reporting throughout prevents the large synthetic OMI Health
corpus from obscuring the smaller, clinically closer ACI-Bench and PriMock57
datasets.

Our contributions are:
\begin{itemize}
    \item A source-aware evaluation benchmark for SOAP note generation that
          preserves dataset identity and reports source-macro averages as the
          primary criterion.
    \item A controlled $2{\times}2$ (reasoning $\times$ RAG) comparison of
          three frontier models across 326 test examples.
    \item A dual LLM-as-a-judge evaluation that scores candidate notes for
          faithfulness, clinical coverage, safety, and SOAP structure using
          the dialogue, reference note, and candidate note jointly.
    \item Empirical evidence---consistent across seven automatic metrics and
          two independent judges---that reasoning capability does not reliably
          improve SOAP note generation and can substantially harm it.
\end{itemize}

\section{Related Work}
Work on clinical note generation from dialogue has focused primarily on benchmark creation and generation pipelines.
ACI-Bench introduced a benchmark of role-played clinical encounters with structured visit notes \citep{yim2023aci}, while PriMock57 released mock primary-care consultations with transcripts and notes to enable documentation research without PHI constraints \citep{papadopouloskorfiatis2022primock57}.
The MEDIQA-Chat 2023 shared task standardized evaluation across note generation subtasks and established few-shot GPT-4 as a strong baseline \citep{benabacha2023mediqa}.
More recent systems move toward modular pipelines: ClinicSum combines retrieval-based filtering with fine-tuned generation to reduce hallucination \citep{neupane2025clinicsum}, and SpecialtyScribe introduces a three-component architecture for oncology that matches frontier model performance at a fraction of the cost \citep{goyal2025specialtyscribe}.
These studies establish that LLMs and retrieval-augmented pipelines can be useful for clinical documentation, but they do not isolate whether explicit reasoning modes improve the generation of structured SOAP notes.
Our work differs from all of these in that it is not a new generation architecture; it is a controlled evaluation of whether reasoning and retrieval independently and jointly improve SOAP note quality.

Clinical RAG systems have shown that retrieval can improve structured documentation, especially when evidence is fragmented.
ClinicSum uses retrieval-based filtering before generation \citep{neupane2025clinicsum}, MedRAG benchmarks multiple corpora against a medical QA framework and reports up to 18\% improvement over chain-of-thought prompting \citep{xiong2024medrag}, and CLI-RAG proposes hierarchical retrieval over heterogeneous EHR note types for progress-note synthesis \citep{keerthana2025clirag}.
These approaches motivate our retrieval condition but treat RAG as a standalone intervention without isolating reasoning mode as an experimental variable.
Thus, while prior work has studied RAG for clinical note and EHR generation, the specific impact of provider-native reasoning on generated SOAP-note quality remains underexplored.

Recent medical reasoning benchmarks report encouraging results on diagnostic and treatment tasks \citep{qiu2025medr}, and o1-style models show gains on reasoning-intensive medical QA \citep{xie2024o1, chen2024huatuogpt}.
However, the evidence for clinical \emph{text} tasks is less straightforward: a large-scale study across 95 LLMs and 87 clinical text understanding tasks found that 86.3\% of models suffered consistent performance degradation with chain-of-thought, attributing the failure to the noisy and fragmented nature of clinical text \citep{wu2025cotfails}.
Evaluation-oriented work such as MEDIVAL examines reasoning-enhanced \emph{judging} of documentation rather than reasoning-enhanced \emph{generation} of it \citep{choi2025medival}, leaving the generation direction open.
The closest work to ours integrates CoT and RAG for rare disease \emph{diagnosis} from clinical notes, finding that architecture order matters depending on note quality \citep{wang2025cotrag}, but does not examine the interaction for structured SOAP note \emph{generation}, where the output is fixed-format and fidelity-sensitive.
A scoping review of 67 RAG studies in medical domains confirms that only 26 included explicit reasoning support, with most treating RAG and reasoning as independently deployed rather than jointly evaluated \citep{neha2025ragscoping}.

Finally, automatic metrics are known to correlate weakly with expert judgment for clinical text \citep{moramarco2022human}, motivating the use of LLM judges as a complementary evaluation signal.
Rubric-based LLM evaluation achieves clinician-level inter-rater reliability at substantially lower cost \citep{shah2025rubrics}, and multi-judge setups help surface calibration differences across judge models \citep{brake2024comparing}.
Our dual-judge design---using GPT-5.4 mini and Gemma-4-E4B with a shared rubric emphasizing faithfulness, clinical coverage, safety, and SOAP structure---follows this paradigm and lets us assess whether qualitative rankings hold across judge families.

\section{Method}

\subsection{Source-aware benchmark}

We normalise three publicly available datasets into a unified, source-aware
benchmark while retaining a \texttt{source\_dataset} field for every example.
OMI Health contributes 10{,}000 synthetic dialogue--SOAP pairs based on
NoteChat-style clinical-note-conditioned conversations, ACI-Bench contributes
112 role-played clinical encounters, and PriMock57 contributes 57 mock
primary-care consultations \citep{omihealthdataset, wang2024notechat,
yim2023aci, papadopouloskorfiatis2022primock57}.
The combined corpus contains 10{,}169 examples split into 9{,}324 train,
519 validation, and 326 test items.

Because OMI Health contributes 97\% of examples, pooled means would mask
performance differences on the two smaller, clinically closer datasets.
We therefore adopt \emph{source-macro} averages---the unweighted mean of the three per-source means---as our primary reporting criterion, with pooled means provided as secondary context.

\subsection{Experimental design}

We evaluate each test example under a $2 \times 2$ factorial design over reasoning and RAG (each in $\{\text{off}, \text{on}\}$).

\paragraph{Reasoning.}
Reasoning is toggled exclusively through provider-native mechanisms; no
``let's think step by step'' prompt variants are used, ensuring that any
observed effect is attributable to the model's internal reasoning process
rather than prompt phrasing.
This strict definition restricts the primary comparison to models with
verifiable reasoning-off support: GPT-5.4, DeepSeek-V4-Flash, and
Gemma-4-E4B.

\paragraph{RAG.}
The retrieval setup is deliberately minimal.
For each test example, we retrieve exactly one training example from the \emph{same} source dataset using Chroma with \texttt{text-embedding-3-small} embeddings \citep{chromadb2024, openai2024embeddings}. The retrieved example is presented as a stylistic and structural reference, not as factual content to be copied verbatim. This conservative design isolates the structural guidance that in-context examples can provide without introducing retrieval noise from cross-dataset or off-topic documents.

\paragraph{Evaluation pipeline.} OMI Health, ACI-Bench, and PriMock57 are normalised into a source-aware benchmark. Each test dialogue is passed to GPT-5.4, DeepSeek-V4-Flash, and Gemma-4-E4B under all four reasoning $\times$ RAG conditions. All models are used at their provider-default temperature settings. RAG retrieves one same-source training example per query. Each generated note is scored with seven automatic metrics and two reference-aware LLM judges. Final comparisons use source-macro means so OMI Health does not dominate ACI-Bench and PriMock57.

\subsection{Automatic metrics}

We score each predicted note with ROUGE-1, ROUGE-2, ROUGE-L, METEOR,
chrF++, BERTScore F1, and ClinicalBERT F1 \citep{lin2004rouge,
banerjee2005meteor, popovic2017chrf, zhang2020bertscore,
alsentzer2019clinicalbert}.
For example $i$, model $m$, and condition $c$, the composite score is:
\begin{equation}
s_{i,m,c} = \frac{1}{7}\sum_{k=1}^{7} \mathrm{metric}_{k}(i,m,c).
\end{equation}
The composite aggregates lexical and semantic similarity signals, reducing the
influence of any single metric.

\subsection{LLM-as-a-judge evaluation}

To complement overlap metrics, GPT-5.4 mini and Gemma-4-E4B judge each saved
generation against the dialogue and gold SOAP note on a 1--5 rubric covering
faithfulness, clinical coverage, hallucination, safety, and SOAP organisation.
Using two judge families helps separate robust findings from judge-specific
calibration.

The 326-example test split contains 250 OMI Health, 66 ACI-Bench, and 10
PriMock57 examples. All four conditions use the same examples, so reasoning
and RAG effects are paired within-example comparisons.

\section{Results}

\subsection{Automatic metrics}

Table~\ref{tab:overall} reports source-macro composite scores for all
model--condition pairs.

\paragraph{Without reasoning, GPT-5.4 leads.}
GPT-5.4 achieves the highest source-macro composite without reasoning: 0.515
direct and 0.527 with RAG, the best overall configuration.

\begin{table}[t]
\centering
\scriptsize
\caption{Overall performance under automatic metrics. Rankings use the source-macro composite so each dataset contributes equally; baseline refers to no reasoning + no RAG.}
\label{tab:overall}
\begin{tabular}{lll}
\toprule
Run & Model & Macro mean [95\% CI] \\
\midrule
Reasoning + RAG & DeepSeek-V4-Flash & \textbf{0.512} [0.502, 0.522] \\
Reasoning + RAG & Gemma-4-E4B & 0.506 [0.496, 0.515] \\
Reasoning + RAG & GPT-5.4 & 0.484 [0.451, 0.512] \\
Reasoning & DeepSeek-V4-Flash & \textbf{0.500} [0.490, 0.509] \\
Reasoning & Gemma-4-E4B & 0.494 [0.485, 0.504] \\
Reasoning & GPT-5.4 & 0.454 [0.419, 0.490] \\
RAG & GPT-5.4 & \textbf{0.527} [0.518, 0.535] \\
RAG & DeepSeek-V4-Flash & 0.521 [0.507, 0.534] \\
RAG & Gemma-4-E4B & 0.517 [0.506, 0.527] \\
baseline & GPT-5.4 & \textbf{0.515} [0.507, 0.524] \\
baseline & DeepSeek-V4-Flash & 0.506 [0.497, 0.514] \\
baseline & Gemma-4-E4B & 0.498 [0.488, 0.507] \\
\bottomrule
\end{tabular}
\end{table}

\paragraph{With reasoning, DeepSeek-V4-Flash leads.}
Once reasoning is enabled, DeepSeek-V4-Flash is strongest, reaching 0.500
direct and 0.512 with RAG.

\paragraph{Reasoning imposes a consistent penalty on GPT-5.4.}
Enabling reasoning reduces GPT-5.4's composite score by 0.063 on ACI-Bench,
0.033 on OMI Health, and 0.088 on PriMock57 in the direct setting.
These degradations replicate across all three sources and corroborate evidence
that explicit reasoning can hurt clinical text tasks \citep{wu2025cotfails}.

\paragraph{RAG is helpful but secondary and uneven.}
RAG yields positive but smaller and source-dependent gains; for example, on
OMI Health without reasoning it adds +0.026 for DeepSeek-V4-Flash and +0.033
for Gemma-4-E4B. Table~\ref{tab:deltas} summarises representative deltas.

\begin{table}[t]
\centering
\scriptsize
\caption{Representative per-source automatic-metric deltas. Positive values indicate improvement relative to the matched control.}
\label{tab:deltas}
\resizebox{\columnwidth}{!}{%
\begin{tabular}{llllr}
\toprule
Operation & Model & Source & Setting & $\Delta$ \\
\midrule
Reasoning & GPT-5.4 & ACI & direct & $-0.063$ \\
Reasoning & DeepSeek-V4-Flash & ACI & direct & $-0.018$ \\
Reasoning & Gemma-4-E4B & ACI & direct & $-0.005$ \\
Reasoning & GPT-5.4 & OMI & direct & $-0.033$ \\
Reasoning & DeepSeek-V4-Flash & OMI & direct & $-0.011$ \\
Reasoning & Gemma-4-E4B & OMI & direct & $\mathbf{+0.002}$ \\
Reasoning & GPT-5.4 & PriMock57 & direct & $-0.088$ \\
Reasoning & DeepSeek-V4-Flash & PriMock57 & direct & $\mathbf{+0.014}$ \\
Reasoning & Gemma-4-E4B & PriMock57 & direct & $-0.008$ \\
\addlinespace
RAG & GPT-5.4 & OMI & no reasoning & $+0.006$ \\
RAG & DeepSeek-V4-Flash & OMI & no reasoning & $+0.026$ \\
RAG & Gemma-4-E4B & OMI & no reasoning & $\mathbf{+0.033}$ \\
RAG & GPT-5.4 & PriMock57 & reasoning & $\mathbf{+0.063}$ \\
RAG & DeepSeek-V4-Flash & PriMock57 & reasoning & $+0.001$ \\
RAG & Gemma-4-E4B & PriMock57 & reasoning & $+0.007$ \\
RAG & GPT-5.4 & PriMock57 & no reasoning & $\mathbf{+0.024}$ \\
RAG & DeepSeek-V4-Flash & PriMock57 & no reasoning & $+0.015$ \\
RAG & Gemma-4-E4B & PriMock57 & no reasoning & $+0.017$ \\
\end{tabular}%
}
\end{table}

Full per-metric tables and source decompositions are provided in
Tables~\ref{tab:detailed-automatic-metrics}--\ref{tab:detailed-automatic-metrics-reasoning-rag}.

\subsection{LLM-as-a-judge results}

Table~\ref{tab:judge-detailed} gives full judge scores. Despite scale
differences, both judges rank a non-reasoning GPT-5.4 variant first overall
and DeepSeek-V4-Flash first among reasoning-enabled configurations. They also
agree that reasoning strongly penalises GPT-5.4: by 0.824 (direct) and 0.488
(RAG) under GPT-5.4 mini Judge, and by 1.082 and 0.656 under Gemma-4-E4B Judge.
RAG effects are smaller and judge-dependent, leaving the central conclusion
unchanged.

\section{Discussion}

The empirical pattern across both automatic metrics and LLM judges is
consistent and interpretable.
Structured SOAP note generation appears to reward grounded compression and
format adherence over open-ended deliberation: a model that ``reasons well''
on diagnostic benchmarks does not automatically produce better SOAP notes
from dialogue.
This gap is not attributable to any single dataset or evaluation metric; it
replicates across three heterogeneous sources and two independent judge families.

\section{Conclusion}

We presented a source-aware evaluation of frontier reasoning-capable models
for SOAP note generation under a controlled $2{\times}2$ reasoning $\times$
RAG design.
Across three datasets, seven automatic metrics, and two independent LLM
judges, provider-native reasoning did not consistently improve note quality
and often reduced it substantially---especially for GPT-5.4.
Same-source RAG provided modest, model-dependent gains that were consistently
smaller than the quality gap introduced by enabling reasoning.
DeepSeek-V4-Flash was the most robust model once reasoning was turned on, but
even its best reasoning configuration did not match the top non-reasoning
GPT-5.4 result.

The practical implication is direct: reasoning-enabled LLMs should be
evaluated on clinical documentation tasks in their target configuration,
not assumed to transfer gains from general or diagnostic reasoning benchmarks.
More broadly, structured, fidelity-sensitive generation tasks may constitute
a qualitatively different challenge from the open-ended reasoning tasks where
thinking modes were developed---and they deserve purpose-built evaluation to
reflect that distinction.

\clearpage
\section{Limitations}

Several limitations constrain the conclusions.
First, OMI Health dominates the corpus by volume, so source-macro reporting
is essential; pooled means would overweight the synthetic distribution.
Second, ACI-Bench and PriMock57 are small, making per-source estimates
noisier and per-example effects hard to generalise.
Third, LLM judges are not clinician raters.
They inspect the dialogue, reference note, and candidate note jointly, which
provides a richer signal than overlap metrics alone, but they may encode
model-family preferences and their calibration clearly differs (as the scale
gap between the two judges illustrates).
Fourth, the reference-aware rubric may partly reward paraphrastic similarity
to the gold note's phrasing and structure rather than independent clinical
quality.
Fifth, provider-native reasoning controls are not fully standardised across
vendors, which is why strict reasoning-off support was required as an
inclusion criterion.
This also limited the model set: API budgets and the cost of running all
reasoning $\times$ RAG conditions restricted the study to three frontier
models and two LLM judges, so broader evaluation across additional open,
commercial, smaller, and specialty-tuned clinical models is needed.
Sixth, our analysis is note-level rather than section-level.
We have not yet decomposed performance by SOAP section, so it remains unclear
whether reasoning harms or helps Subjective, Objective, Assessment, and Plan
content in the same way.

These limitations point to productive directions for follow-up work: larger
clinically realistic test sets, expert physician evaluation, expanded model
coverage under larger API budgets, section-wise SOAP analysis, and
model-agnostic reasoning interventions that allow broader comparison.

\clearpage
\bibliography{references}

\clearpage
\onecolumn

\appendix

\section{Additional Figures}

\begin{figure}[H]
\centering
\includegraphics[width=0.50\textwidth]{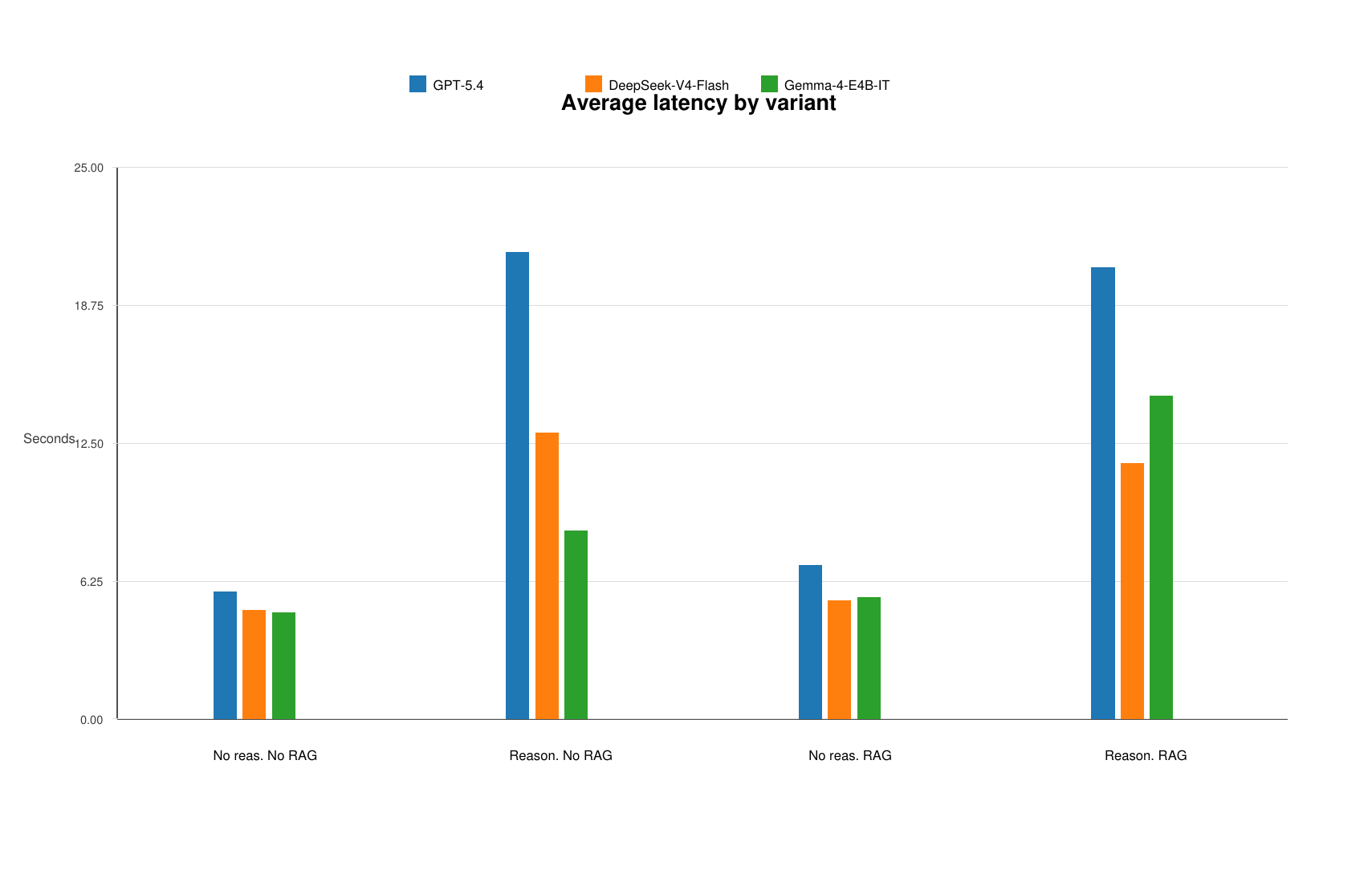}
\caption{Latency analysis for the saved provider reasoning run.
Provider-native reasoning increases latency for several model--condition pairs;
quality comparisons should therefore be read alongside efficiency costs.}
\label{fig:latency}
\end{figure}

\begin{figure}[H]
\centering
\includegraphics[width=0.50\textwidth]{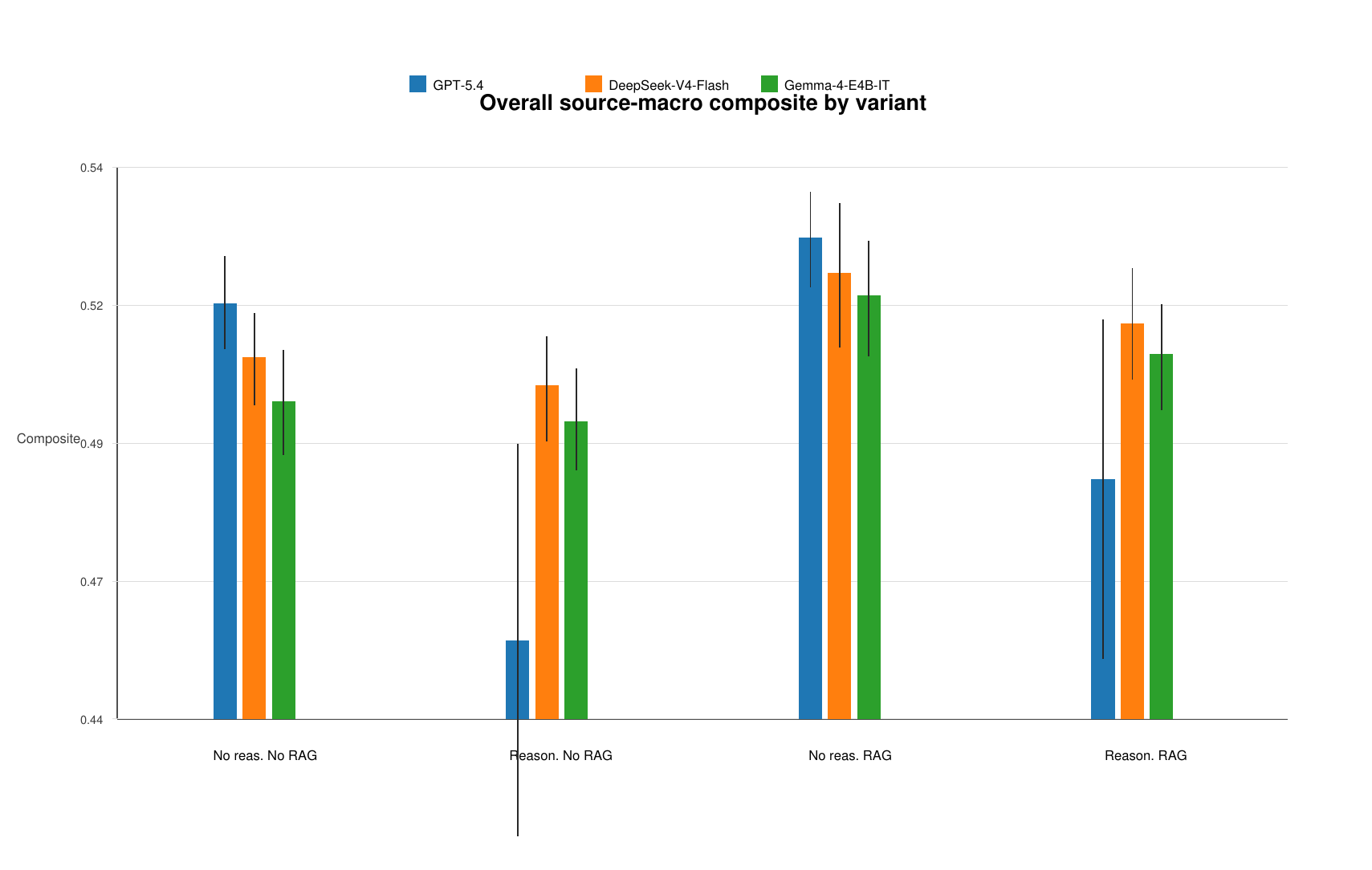}
\caption{Source-macro composite score by model and test variant. Error bars are 95\% bootstrap confidence
intervals.}
\label{fig:overall-model-performance}
\end{figure}

\begin{figure}[H]
\centering
\includegraphics[height=0.50\textwidth]{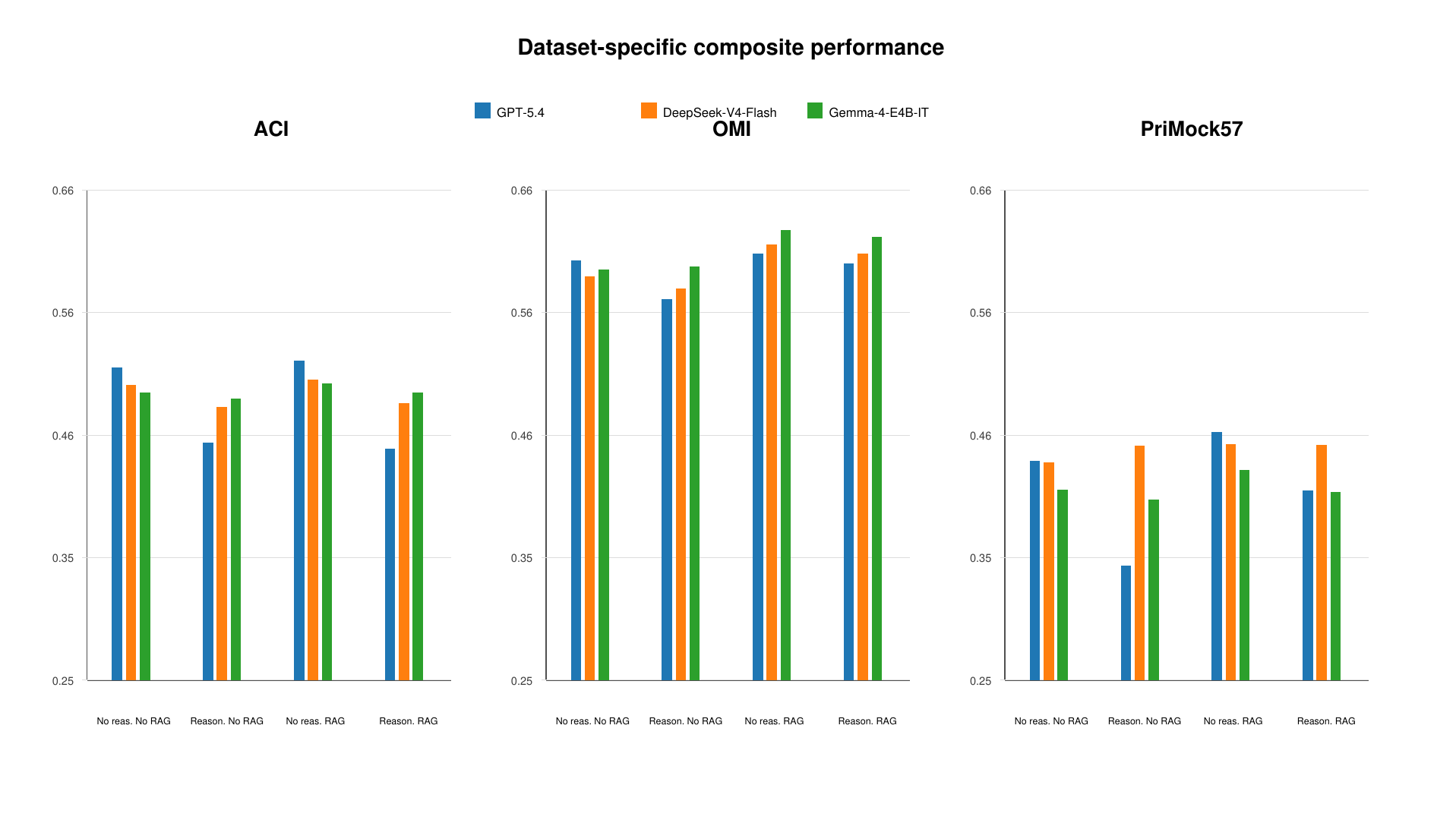}
\caption{Dataset-specific source means for the automatic-metric composite.
The source-aware view exposes differences hidden by pooled averages, including
the stronger OMI Health RAG gains relative to PriMock57 for Gemma-4-E4B.}
\label{fig:dataset-specific-performance}
\end{figure}

\begin{figure}[H]
\centering
\includegraphics[width=0.50\textwidth]{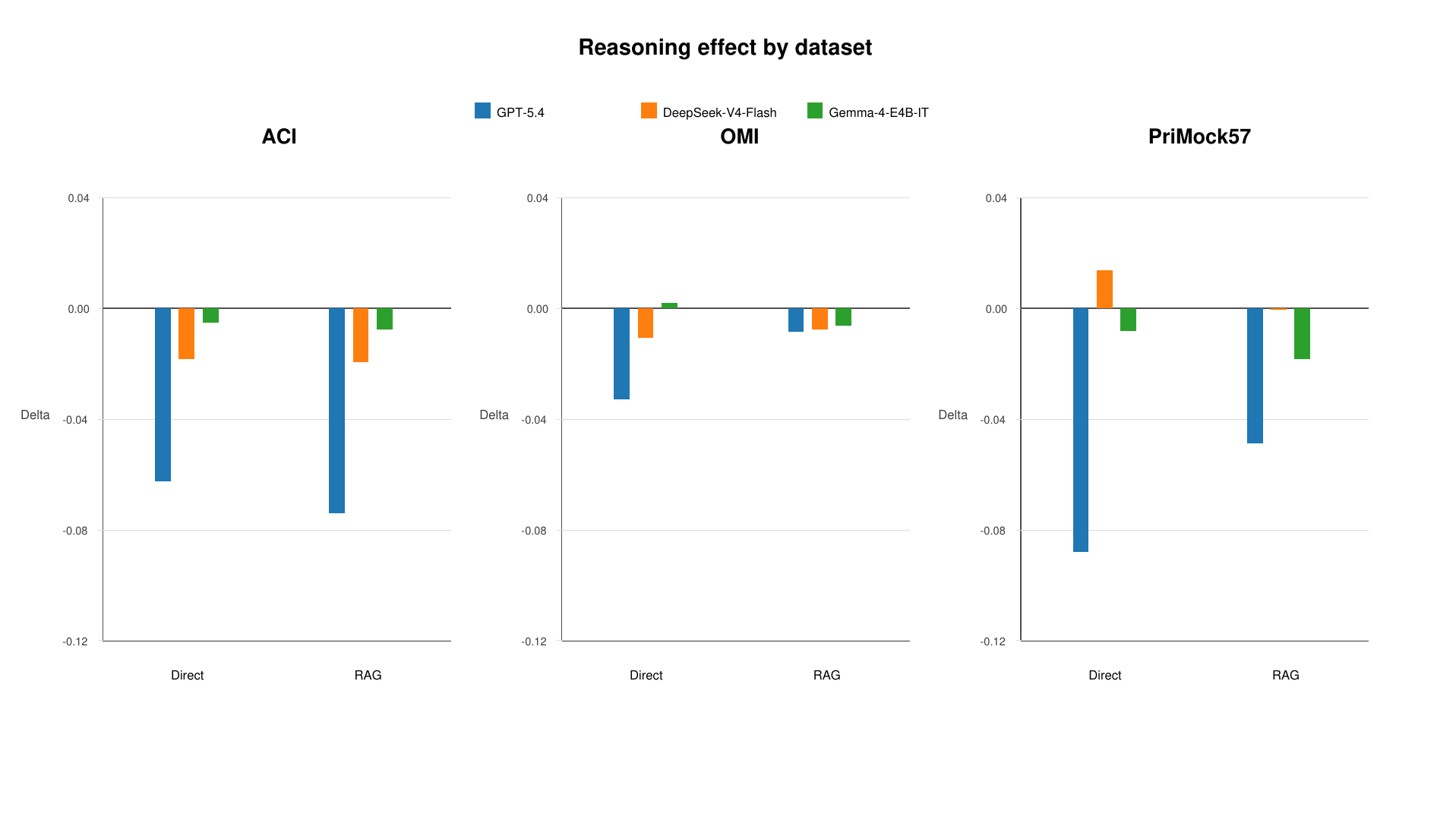}
\caption{Within-example reasoning effects by source dataset and prompting mode.
Positive values indicate improvement after enabling provider-native reasoning
relative to the matched no-reasoning control.}
\label{fig:reasoning-effect}
\end{figure}

\begin{figure}[H]
\centering
\includegraphics[width=0.50\textwidth]{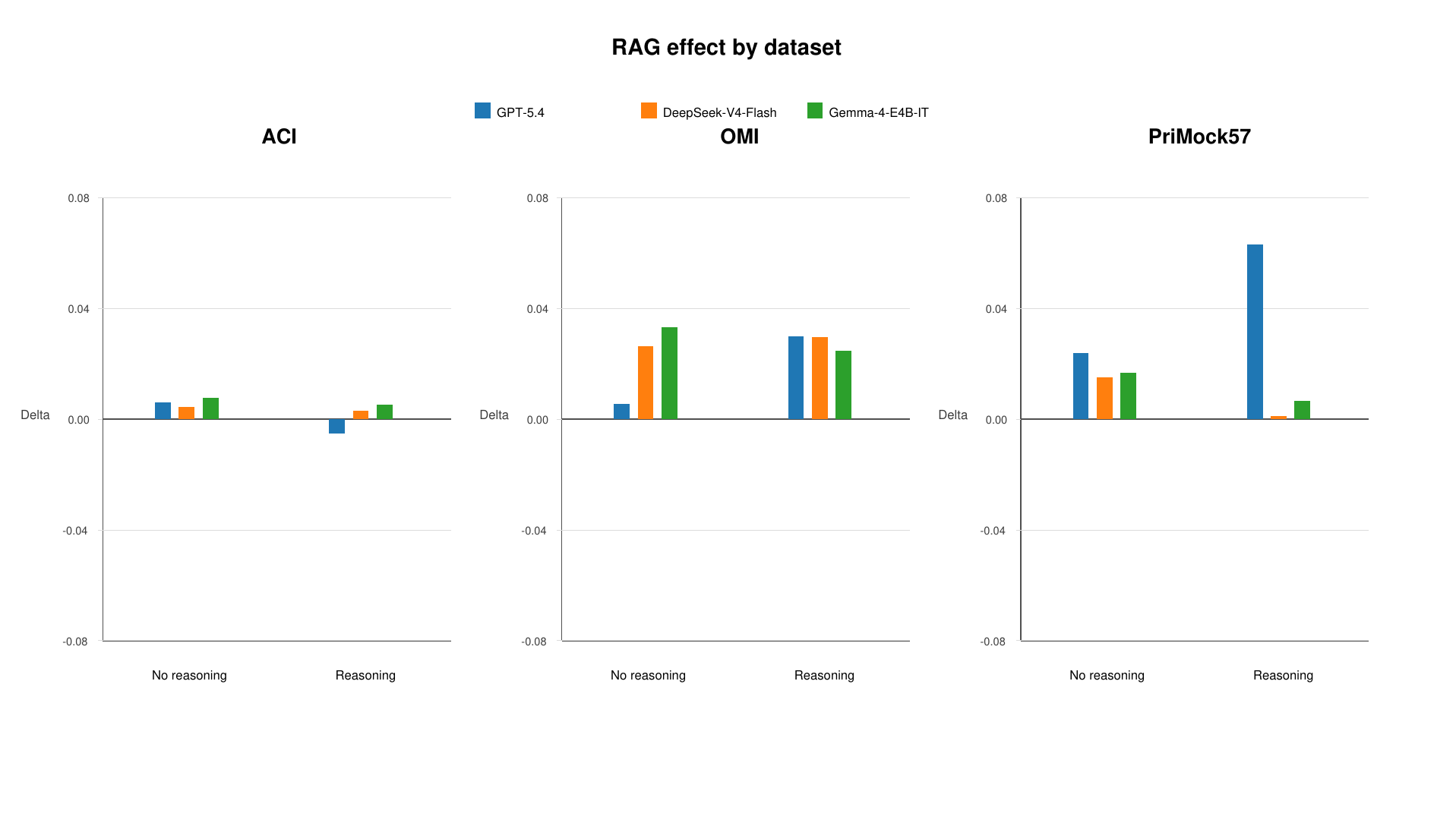}
\caption{Within-example RAG effects by source dataset and reasoning condition.
Positive values indicate improvement after adding the same-source retrieved
training example.}
\label{fig:rag-effect}
\end{figure}

\begin{figure}[H]
\centering
\includegraphics[width=0.50\textwidth]{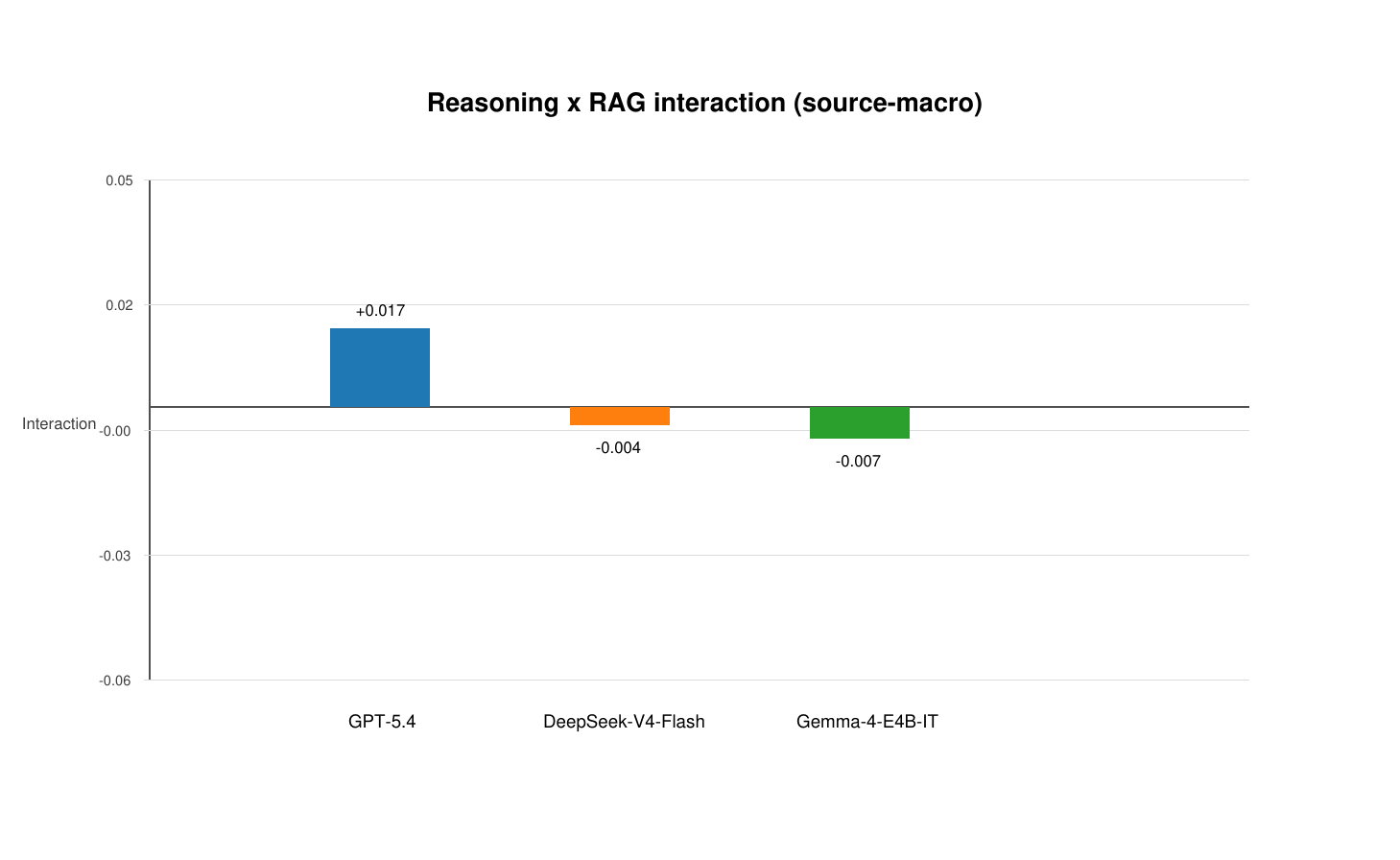}
\caption{Reasoning--RAG interaction effects on the automatic-metric composite.
The interaction is model- and source-dependent, not a uniform additive gain.}
\label{fig:interaction-effects}
\end{figure}

\begin{figure}[H]
\centering
\includegraphics[width=0.50\textwidth]{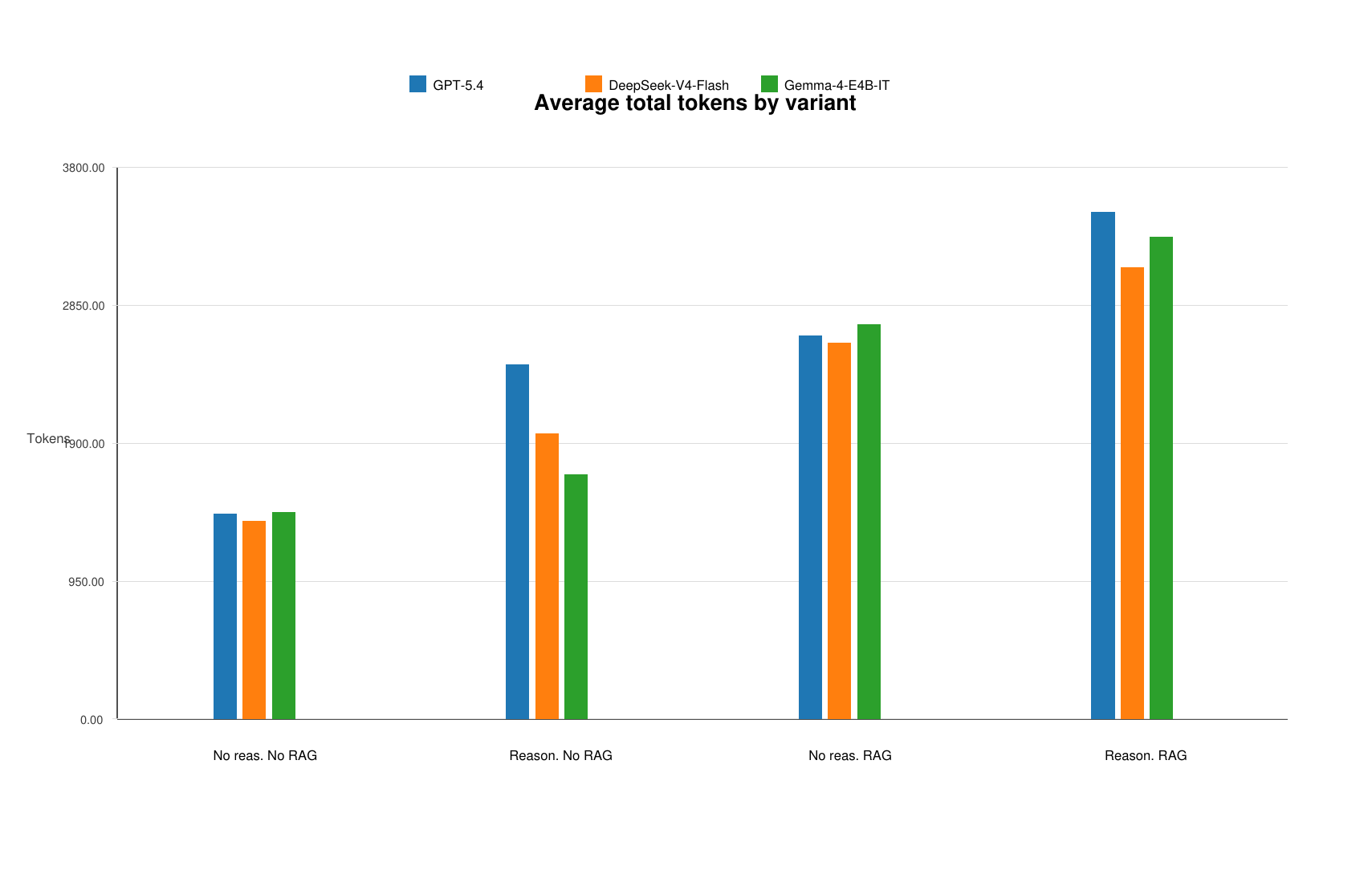}
\caption{Token efficiency analysis for the saved provider reasoning run.
Reasoning substantially increases token use, especially for GPT-5.4, without
a corresponding gain in composite score.}
\label{fig:token-efficiency}
\end{figure}

\clearpage
\section{Additional Tables}

\begin{table}[H]
\centering
\scriptsize
\setlength{\tabcolsep}{4pt}
\caption{Detailed source-macro automatic metric results for every primary model and test variant. Each cell reports mean [95\% bootstrap CI]. CIs resample examples within each source dataset and then average the three source means.}
\label{tab:detailed-automatic-metrics}
\begin{tabular}{llrr}
\toprule
\multicolumn{4}{l}{\textbf{No reasoning + no RAG}} \\
Model & Metric & Source-macro mean [95\% CI] & N \\
\midrule
GPT-5.4 & Composite & \textbf{0.515} [0.507, 0.524] & 326 \\
 & ROUGE-1 & \textbf{0.523} [0.511, 0.537] & 326 \\
 & ROUGE-2 & \textbf{0.222} [0.211, 0.234] & 326 \\
 & ROUGE-L & 0.319 [0.306, 0.331] & 326 \\
 & METEOR & \textbf{0.388} [0.371, 0.402] & 326 \\
 & chrF++ & \textbf{0.456} [0.445, 0.468] & 326 \\
 & BERTScore F1 & 0.872 [0.869, 0.876] & 326 \\
 & ClinicalBERT F1 & \textbf{0.827} [0.823, 0.832] & 326 \\
\addlinespace
DeepSeek-V4-Flash & Composite & 0.506 [0.497, 0.514] & 326 \\
 & ROUGE-1 & 0.511 [0.498, 0.523] & 326 \\
 & ROUGE-2 & 0.214 [0.203, 0.225] & 326 \\
 & ROUGE-L & \textbf{0.321} [0.307, 0.334] & 326 \\
 & METEOR & 0.365 [0.347, 0.380] & 326 \\
 & chrF++ & 0.430 [0.418, 0.441] & 326 \\
 & BERTScore F1 & \textbf{0.874} [0.870, 0.877] & 326 \\
 & ClinicalBERT F1 & 0.826 [0.821, 0.830] & 326 \\
\addlinespace
Gemma-4-E4B & Composite & 0.498 [0.488, 0.507] & 326 \\
 & ROUGE-1 & 0.500 [0.488, 0.513] & 326 \\
 & ROUGE-2 & 0.210 [0.199, 0.222] & 326 \\
 & ROUGE-L & 0.309 [0.298, 0.320] & 326 \\
 & METEOR & 0.353 [0.334, 0.370] & 326 \\
 & chrF++ & 0.417 [0.402, 0.432] & 326 \\
 & BERTScore F1 & 0.872 [0.869, 0.875] & 326 \\
 & ClinicalBERT F1 & 0.821 [0.816, 0.827] & 326 \\
\bottomrule
\end{tabular}
\end{table}

\begin{table}[H]
\centering
\scriptsize
\setlength{\tabcolsep}{4pt}
\caption{Detailed source-macro automatic metric results for Reasoning + no RAG.}
\label{tab:detailed-automatic-metrics-reasoning-no-rag}
\begin{tabular}{llrr}
\toprule
\multicolumn{4}{l}{\textbf{Reasoning + no RAG}} \\
Model & Metric & Source-macro mean [95\% CI] & N \\
\midrule
GPT-5.4 & Composite & 0.454 [0.419, 0.490] & 326 \\
 & ROUGE-1 & 0.422 [0.363, 0.481] & 326 \\
 & ROUGE-2 & 0.183 [0.159, 0.207] & 326 \\
 & ROUGE-L & 0.268 [0.236, 0.300] & 326 \\
 & METEOR & 0.305 [0.256, 0.354] & 326 \\
 & chrF++ & 0.353 [0.297, 0.408] & 326 \\
 & BERTScore F1 & 0.860 [0.851, 0.869] & 326 \\
 & ClinicalBERT F1 & 0.789 [0.767, 0.810] & 326 \\
\addlinespace
DeepSeek-V4-Flash & Composite & \textbf{0.500} [0.490, 0.509] & 326 \\
 & ROUGE-1 & \textbf{0.506} [0.492, 0.521] & 326 \\
 & ROUGE-2 & \textbf{0.210} [0.197, 0.223] & 326 \\
 & ROUGE-L & \textbf{0.314} [0.301, 0.327] & 326 \\
 & METEOR & \textbf{0.354} [0.336, 0.370] & 326 \\
 & chrF++ & \textbf{0.421} [0.408, 0.433] & 326 \\
 & BERTScore F1 & \textbf{0.873} [0.870, 0.876] & 326 \\
 & ClinicalBERT F1 & \textbf{0.824} [0.819, 0.829] & 326 \\
\addlinespace
Gemma-4-E4B & Composite & 0.494 [0.485, 0.504] & 326 \\
 & ROUGE-1 & 0.498 [0.484, 0.512] & 326 \\
 & ROUGE-2 & 0.208 [0.196, 0.219] & 326 \\
 & ROUGE-L & 0.306 [0.294, 0.318] & 326 \\
 & METEOR & 0.340 [0.324, 0.356] & 326 \\
 & chrF++ & 0.413 [0.402, 0.426] & 326 \\
 & BERTScore F1 & \textbf{0.873} [0.870, 0.876] & 326 \\
 & ClinicalBERT F1 & 0.820 [0.815, 0.825] & 326 \\
\bottomrule
\end{tabular}
\end{table}

\begin{table}[H]
\centering
\scriptsize
\setlength{\tabcolsep}{4pt}
\caption{Detailed source-macro automatic metric results for No reasoning + RAG.}
\label{tab:detailed-automatic-metrics-no-reasoning-rag}
\begin{tabular}{llrr}
\toprule
\multicolumn{4}{l}{\textbf{No reasoning + RAG}} \\
Model & Metric & Source-macro mean [95\% CI] & N \\
\midrule
GPT-5.4 & Composite & \textbf{0.527} [0.518, 0.535] & 326 \\
 & ROUGE-1 & \textbf{0.540} [0.529, 0.552] & 326 \\
 & ROUGE-2 & 0.230 [0.218, 0.242] & 326 \\
 & ROUGE-L & 0.332 [0.318, 0.347] & 326 \\
 & METEOR & \textbf{0.404} [0.385, 0.420] & 326 \\
 & chrF++ & \textbf{0.472} [0.461, 0.483] & 326 \\
 & BERTScore F1 & 0.876 [0.873, 0.880] & 326 \\
 & ClinicalBERT F1 & \textbf{0.835} [0.831, 0.838] & 326 \\
\addlinespace
DeepSeek-V4-Flash & Composite & 0.521 [0.507, 0.534] & 326 \\
 & ROUGE-1 & 0.530 [0.512, 0.547] & 326 \\
 & ROUGE-2 & \textbf{0.234} [0.218, 0.249] & 326 \\
 & ROUGE-L & \textbf{0.343} [0.327, 0.359] & 326 \\
 & METEOR & 0.378 [0.355, 0.401] & 326 \\
 & chrF++ & 0.450 [0.432, 0.466] & 326 \\
 & BERTScore F1 & \textbf{0.879} [0.875, 0.883] & 326 \\
 & ClinicalBERT F1 & 0.832 [0.824, 0.839] & 326 \\
\addlinespace
Gemma-4-E4B & Composite & 0.517 [0.506, 0.527] & 326 \\
 & ROUGE-1 & 0.530 [0.516, 0.544] & 326 \\
 & ROUGE-2 & 0.233 [0.222, 0.244] & 326 \\
 & ROUGE-L & 0.330 [0.317, 0.342] & 326 \\
 & METEOR & 0.371 [0.350, 0.391] & 326 \\
 & chrF++ & 0.445 [0.425, 0.462] & 326 \\
 & BERTScore F1 & 0.878 [0.875, 0.882] & 326 \\
 & ClinicalBERT F1 & 0.830 [0.825, 0.835] & 326 \\
\bottomrule
\end{tabular}
\end{table}

\begin{table}[H]
\centering
\scriptsize
\setlength{\tabcolsep}{4pt}
\caption{Detailed source-macro automatic metric results for Reasoning + RAG.}
\label{tab:detailed-automatic-metrics-reasoning-rag}
\begin{tabular}{llrr}
\toprule
\multicolumn{4}{l}{\textbf{Reasoning + RAG}} \\
Model & Metric & Source-macro mean [95\% CI] & N \\
\midrule
GPT-5.4 & Composite & 0.484 [0.451, 0.512] & 326 \\
 & ROUGE-1 & 0.470 [0.415, 0.517] & 326 \\
 & ROUGE-2 & 0.197 [0.174, 0.218] & 326 \\
 & ROUGE-L & 0.290 [0.260, 0.316] & 326 \\
 & METEOR & 0.349 [0.303, 0.390] & 326 \\
 & chrF++ & 0.405 [0.354, 0.450] & 326 \\
 & BERTScore F1 & 0.866 [0.858, 0.873] & 326 \\
 & ClinicalBERT F1 & 0.808 [0.787, 0.827] & 326 \\
\addlinespace
DeepSeek-V4-Flash & Composite & \textbf{0.512} [0.502, 0.522] & 326 \\
 & ROUGE-1 & \textbf{0.518} [0.504, 0.533] & 326 \\
 & ROUGE-2 & \textbf{0.226} [0.212, 0.239] & 326 \\
 & ROUGE-L & \textbf{0.332} [0.317, 0.347] & 326 \\
 & METEOR & \textbf{0.368} [0.350, 0.384] & 326 \\
 & chrF++ & \textbf{0.434} [0.417, 0.449] & 326 \\
 & BERTScore F1 & 0.875 [0.872, 0.879] & 326 \\
 & ClinicalBERT F1 & \textbf{0.828} [0.824, 0.833] & 326 \\
\addlinespace
Gemma-4-E4B & Composite & 0.506 [0.496, 0.515] & 326 \\
 & ROUGE-1 & 0.516 [0.502, 0.530] & 326 \\
 & ROUGE-2 & 0.220 [0.208, 0.233] & 326 \\
 & ROUGE-L & 0.322 [0.309, 0.337] & 326 \\
 & METEOR & 0.353 [0.334, 0.369] & 326 \\
 & chrF++ & 0.429 [0.412, 0.444] & 326 \\
 & BERTScore F1 & \textbf{0.877} [0.874, 0.879] & 326 \\
 & ClinicalBERT F1 & 0.826 [0.821, 0.830] & 326 \\
\bottomrule
\end{tabular}
\end{table}

\begin{table*}[t]
\centering
\scriptsize
\caption{Detailed LLM-judge results. `ACI`, `OMI`, and `Primock` are per-source mean judge scores on the 1--5 rubric; `Macro` is their unweighted average.}
\label{tab:judge-detailed}
\begin{tabular}{lllrrrr}
\toprule
Judge & Model & Condition & ACI & OMI & Pri & Macro \\
\midrule
GPT-5.4 mini & GPT-5.4 & No reasoning + no RAG & \textbf{4.121} & 4.080 & \textbf{4.100} & \textbf{4.100} \\
GPT-5.4 mini & DeepSeek-V4-Flash & No reasoning + no RAG & 4.106 & 3.872 & 4.000 & 3.993 \\
GPT-5.4 mini & Gemma-4-E4B & No reasoning + no RAG & 3.970 & 3.880 & 4.000 & 3.950 \\
GPT-5.4 mini & GPT-5.4 & No reasoning + RAG & \textbf{4.121} & \textbf{4.128} & 4.000 & 4.083 \\
GPT-5.4 mini & DeepSeek-V4-Flash & No reasoning + RAG & 3.955 & 4.000 & 3.900 & 3.952 \\
GPT-5.4 mini & Gemma-4-E4B & No reasoning + RAG & 3.924 & 3.988 & 3.857 & 3.923 \\
GPT-5.4 mini & DeepSeek-V4-Flash & Reasoning + no RAG & 4.015 & 3.888 & 4.000 & 3.968 \\
GPT-5.4 mini & Gemma-4-E4B & Reasoning + no RAG & 3.985 & 3.896 & 3.900 & 3.927 \\
GPT-5.4 mini & GPT-5.4 & Reasoning + no RAG & 3.561 & 3.768 & 2.500 & 3.276 \\
GPT-5.4 mini & DeepSeek-V4-Flash & Reasoning + RAG & 3.970 & 3.996 & 3.900 & 3.955 \\
GPT-5.4 mini & Gemma-4-E4B & Reasoning + RAG & 3.939 & 3.980 & 3.857 & 3.926 \\
GPT-5.4 mini & GPT-5.4 & Reasoning + RAG & 3.409 & 4.076 & 3.300 & 3.595 \\
\midrule
Gemma-4-E4B & GPT-5.4 & No reasoning + no RAG & 4.894 & 4.216 & \textbf{5.000} & 4.703 \\
Gemma-4-E4B & DeepSeek-V4-Flash & No reasoning + no RAG & 4.848 & 3.836 & 4.800 & 4.495 \\
Gemma-4-E4B & Gemma-4-E4B & No reasoning + no RAG & 4.652 & 3.688 & 4.800 & 4.380 \\
Gemma-4-E4B & GPT-5.4 & No reasoning + RAG & \textbf{4.924} & \textbf{4.532} & \textbf{5.000} & \textbf{4.819} \\
Gemma-4-E4B & DeepSeek-V4-Flash & No reasoning + RAG & 4.636 & 4.336 & 4.800 & 4.591 \\
Gemma-4-E4B & Gemma-4-E4B & No reasoning + RAG & 4.591 & 4.192 & 4.143 & 4.309 \\
Gemma-4-E4B & DeepSeek-V4-Flash & Reasoning + no RAG & 4.879 & 3.892 & 4.900 & 4.557 \\
Gemma-4-E4B & Gemma-4-E4B & Reasoning + no RAG & 4.758 & 3.952 & 4.800 & 4.503 \\
Gemma-4-E4B & GPT-5.4 & Reasoning + no RAG & 3.955 & 3.908 & 3.000 & 3.621 \\
Gemma-4-E4B & DeepSeek-V4-Flash & Reasoning + RAG & 4.727 & 4.360 & \textbf{5.000} & 4.696 \\
Gemma-4-E4B & Gemma-4-E4B & Reasoning + RAG & 4.561 & 4.208 & 4.857 & 4.542 \\
Gemma-4-E4B & GPT-5.4 & Reasoning + RAG & 3.939 & 4.448 & 4.100 & 4.162 \\
\bottomrule
\end{tabular}
\end{table*}

\clearpage

\section{Generation and Judge Prompts}

\begin{tcblisting}{title=Direct SOAP Generation Prompt, colback=white, colframe=black, fonttitle=\bfseries, breakable, listing only, listing options={basicstyle=\ttfamily\small, breaklines=true, breakatwhitespace=false, columns=fullflexible, keepspaces=true, showstringspaces=false}}
Generate a clinical SOAP note from the dialogue below.

Use only information explicitly stated in the dialogue.
Do not infer, guess, add normal exam findings, invent diagnoses, or add treatment details not mentioned.

Return only the final SOAP note.
Do not include any introduction, explanation, markdown fences, bullets, or extra headings.
Use exactly these four section headers, in this exact order, once each:

S:
O:
A:
P:

Formatting requirements:
- Put content on the same line as each header when possible.
- If a section has no supporting information in the dialogue, write `No information provided.`
- Keep the note concise and clinically neutral.
- Preserve uncertainty when the dialogue is uncertain.

Section guidance:
- `S:` symptoms, patient-reported history, concerns, and relevant subjective context.
- `O:` observed findings, exam findings, measurements, test results, and other objective facts.
- `A:` clinician assessment, impression, working diagnosis, or differential if explicitly stated.
- `P:` plan, next steps, treatment, follow-up, referrals, instructions, or monitoring if explicitly stated.

Dialogue:
{{ dialogue_text }}
\end{tcblisting}

\begin{tcblisting}{title=RAG SOAP Generation Prompt, colback=white, colframe=black, fonttitle=\bfseries, breakable, listing only, listing options={basicstyle=\ttfamily\small, breaklines=true, breakatwhitespace=false, columns=fullflexible, keepspaces=true, showstringspaces=false}}
Generate a clinical SOAP note from the target dialogue.

You are given {{ retrieval_k }} retrieved training example(s)
Use the retrieved examples only as demonstrations of note style and organization.
Do not copy facts, diagnoses, plans, or wording from the retrieved examples into the target note unless they are explicitly stated in the target dialogue.

Use only information explicitly stated in the target dialogue.
Do not infer, guess, add normal exam findings, invent diagnoses, or add treatment details not mentioned.

Return only the final SOAP note.
Do not include any introduction, explanation, markdown fences, bullets, or extra headings.
Use exactly these four section headers, in this exact order, once each:

S:
O:
A:
P:

Formatting requirements:
- Put content on the same line as each header when possible.
- If a section has no supporting information in the target dialogue, write `No information provided.`
- Keep the note concise and clinically neutral.
- Preserve uncertainty when the target dialogue is uncertain.

Retrieved training examples:
{% for example in retrieved_examples %}
Example {{ loop.index }} conversation:
{{ example.dialogue_text }}

Example {{ loop.index }} SOAP note:
{{ example.soap_note }}

{% endfor %}
Target dialogue:
{{ dialogue_text }}

\end{tcblisting}

\begin{tcblisting}{title=Clinical Correctness Judge Prompt, colback=white, colframe=black, fonttitle=\bfseries, breakable, listing only, listing options={basicstyle=\ttfamily\small, breaklines=true, breakatwhitespace=false, columns=fullflexible, keepspaces=true, showstringspaces=false}}
You are a blinded clinical documentation evaluator. Score the generated SOAP note for clinical correctness of diagnoses, symptoms, medications, tests, and plans.

Use this 1-5 scale:
1 = unsafe or mostly incorrect
2 = major omissions or unsupported content
3 = mixed quality with notable issues
4 = clinically acceptable with minor issues
5 = excellent, faithful, and complete

Return strict JSON only with keys:
{"score": number, "rationale": string, "hallucinated_claims": [string], "missing_key_facts": [string], "unsafe_recommendations": [string]}

Dialogue:
{{ dialogue_text }}

Reference SOAP note:
{{ reference_note }}

Generated SOAP note:
{{ generated_note }}
\end{tcblisting}

\end{document}